\documentclass{article}

\usepackage{PRIMEarxiv}

\usepackage[utf8]{inputenc} 
\usepackage[T1]{fontenc}    
\usepackage{hyperref}       
\usepackage{url}            
\usepackage{booktabs}       
\usepackage{amsfonts}       
\usepackage{nicefrac}       
\usepackage{microtype}      
\usepackage{lipsum}
\usepackage{fancyhdr}       
\usepackage{graphicx}       
\graphicspath{{media/}}     

\usepackage{rsfso}
\usepackage{amsmath}
\usepackage{graphicx}
\usepackage{subfig}
\usepackage{mwe}

\title{A Novel Loss Function Utilizing Wasserstein Distance to Reduce Subject-Dependent Noise for Generalizable Models in Affective Computing
}

\author{%
  Nibraas A. Khan \\
  Department of Computer Science\\
  Vanderbilt University\\
  Nashville, TN 37235 \\
  \texttt{nibraas.a.khan@vanderbilt.edu} \\
   \And
   Mahrukh Tauseef \\
   Department of Electrical Engineering \\
   Vanderbilt University\\
   Nashville, TN 37235 \\
   \texttt{mahrukh.tauseef@vanderbilt.edu} \\
   \And
   Ritam Ghosh \\
   Department of Electrical Engineering \\
   Vanderbilt University\\
   Nashville, TN 37235 \\
   \texttt{ritam.ghosh@vanderbilt.edu} \\
  \And
  Nilanjan Sarkar \\
  Department of Mechanical Engineering \\
  Vanderbilt University\\
  Nashville, TN 37235 \\
  \texttt{nilanjan.sarkar@vanderbilt.edu} \\
}

\begin{document}
\maketitle

\begin{abstract}
Emotions are an essential part of human behavior that can impact thinking, decision-making, and communication skills. Thus, the ability to accurately monitor and identify emotions can be useful in many human-centered applications such as behavioral training, tracking emotional well-being, and development of human-computer interfaces. The correlation between patterns in physiological data and affective states has allowed for the utilization of deep learning techniques which can accurately detect the affective states of a person. However, the generalisability of existing models is often limited by the subject-dependent noise in the physiological data due to variations in a subject’s reactions to stimuli. Hence, we propose a novel cost function that employs Optimal Transport Theory, specifically Wasserstein Distance, to scale the importance of subject-dependent data such that higher importance is assigned to patterns in data that are common across all participants while decreasing the importance of patterns that result from subject-dependent noise. The performance of the proposed cost function is demonstrated through an autoencoder with a multi-class classifier attached to the latent space and trained simultaneously to detect different affective states. An autoencoder with a state-of-the-art loss function i.e., Mean Squared Error, is used as a baseline for comparison with our model across four different commonly used datasets. Centroid and minimum distance between different classes are used as a metrics to indicate the separation between different classes in the latent space. An average increase of $14.75\%$ and $17.75\%$ (from benchmark to proposed loss function) was found for minimum and centroid euclidean distance respectively over all datasets.
\end{abstract}


\section{Introduction}
Affective state can be defined as the underlying experience of feeling, emotion, or mood \cite{Hogg2007}. Its importance stems from the influence affective states have on a person's thinking skills, decision making, and communication skills as well as mental and physical well-being \cite{YADE}. Accurate monitoring and identification of affective states can lead to important applications in behavioral training, computer-based emotional analysis (e.g., stress detection), and human-computer interfaces that can cater to the emotional needs of an individual \cite{ARYA2021100399,Khateeb}.

Affective state can be measured and monitored in two ways: intrusively and non-intrusively. Intrusive methods measure the concentration of various hormones in the blood stream that can be used to detect an affective state. For instance, cortisol levels produced by the hypothalamic-pituitary-adrenocortical (HPA) axis can be collected through samples of blood, urine, hair or saliva and used to detect stress \cite{Greene2016}. One of the challenges with intrusive measurement is that it is invasive and cannot be used to monitor the affective state in real-time. 

Non-intrusive methods, on the other hand, include the analysis of behavioral or physiological data that can lead to non-invasive real-time detection of affective states. Behavioural actions such as blink rate, facial expression, gesture, speech, and body pose have been commonly used for affective state detection \cite{shu2018review}. However, all these actions can be masked and voluntarily controlled by a subject which reduces its reliability \cite{Khateeb,shu2018review}. Alternatively, strong evidence suggests physiological signals to be more reliable for affective state detection due to its involuntary nature and a strong correlation of patterns in the data with different affective states \cite{Greene2016}. 

Physiological signals are a response to the Autonomic Nervous System (ANS) utilizing both motor and sensory neurons to communicate and operate between the Central Nervous System (CNS) and the various organs or muscles. The ANS is composed of the Sympathetic (SNS) and Parasympathetic (PNS) nervous systems. The sympathetic nervous system prepares the body for emergency action, which results in the “fight or flight” response \cite{Richter2013}. The response leads to an increase in measurable physiological signals, such as heart rate, blood flow, and increased muscle activation which can be mapped to different affective states. Whereas, the parasympathetic nervous system helps sustain homeostasis during rest by decreasing physiological signals and maintaining them in moderate ranges \cite{Glick1965}. Since ANS is involuntarily stimulated, the response cannot be manipulated or masked by an individual. This has led to the popular field of affecting computing to focus on dynamically identifying different affective states using non-invasive wearable sensors that can monitor changes in physiological signals in response to a stimulus.

The ability to map patterns in physiological data with affective states has allowed for utilization of machine learning techniques for the detection of affective states. Research in psychophysiology has led to the compilation of a comprehensive list of physiological data that can be used to monitor affective state. This list includes heart activity (ECG), brain activity (EEG), skin response (EDA), blood pressure variation (PPG), respiratory response, and muscle activity (EMG) \cite{Greene2016}. Several datasets have been compiled by collecting physiological data while inducing an affective state. For instance, one of the state-of-the-art datasets, Wearable Stress and Affect Detection (WESAD) \cite{schmidt2018introducing}, induced amusement by making the subjects watch funny video clips and they induced stress through public speaking and mental arithmetic tasks. Meanwhile, their physiological data i.e., blood volume pulse, ECG, EDA, EMG, respiration, body temperature, and three-axis acceleration was collected and classified simultaneously. Other common datasets like Database for Emotional Analysis using Physiological Signals (DEAP) \cite{koelstra2011deap}, Affect, Personality and Mood Research on Individuals and Groups (AMIGOS) \cite{miranda2018amigos}, and Cognitive Load, Affect, and Stress Recognition (CLAS) \cite{clas} follow suit by using video clips or stress-inducing tasks to collect and classify physiological data for the induced affective states. 


Numerous machine learning techniques have been utilised to detect affective states from physiological data. These approaches include support vector machines (SVM), random forest (RF), k-nearest neighbors (KNN) and Linear Discriminant Analysis (LDA) that need handcrafted features from the pre-processed signal in order to remove noisy data \cite{shu2018review}. However, there is no consensus on the list of features extracted from physiological data that can be accurately mapped to affective states which results in reduced performance \cite{li2020stress}. Alternatively, with the advancement of deep learning, there has been an increasing interest in using deep learning techniques like long short-term memory (LSTM), autoencoders, and convolutional neural networks (CNN) \cite{ARYA2021100399,oskooei2021destress}. These models allow for automatic feature extraction based on the model's ability to automatically comprehend patterns in data with respect to the labels. 

Even though deep learning techniques show promise, the lack of generalisability still exists and contributes to poor performance. This is because the subjects might not exhibit the same physiological response for a stimuli \cite{li2020stress}. Thus, subject dependent noise can lead to poor generalisability of the model. This issue can be resolved by using a loss function that filters out the features of data that are person specific and do not contribute much to affective state detection. In other words, if each subject's data is treated as a distribution, the loss function assigns higher importance to features that are closer in distance to the group distribution (distributions of all subjects) while assigning lower importance to features that are much further apart across from the group distribution. The distance between distributions can be calculated by using the Wasserstein distance \cite{monge1781memoir}.

In this paper, we introduce a novel loss function that accounts for subject dependent noise in the data in order to develop more generalizable models. This is accomplished by training an autoencoder model with a loss function that utilizes Wasserstein Distance to scale the importance of subject dependent patterns to obtain a latent space with reduced dimensions and less noisy subject independent features. The performance of the model is tested on four different datasets (WESAD, AMIGOS, CLAS, and DEAP) using the centroid and minimum distances between classes as metrics.

This paper is structured as follows. Section \ref{lit} presents an overview of existing literature that informed the development of the proposed model. This is followed by the \ref{methods} section that discusses the mathematical and algorithmic details of the model. The results are shown and discussed in Section \ref{results} followed by concluding remarks in Section \ref{conclusion}. 

\section{Background}
\label{lit}

As mentioned before, machine learning techniques for affective state detection using physiological data has been a topic of immense interest in the past few years. Kolodyazhniy et al. \cite{kolodyazhniy2011affective} conducted a notable study in 2011 on the topic by compiling a dataset and training several machine learning models like LDA, Quaratic Discriminant Analyis (QDA), Multilayer Perceptron (MLP), Radial Basis Function (RBF), and KNN. They used a variety of features extracted from the physiological data to discover the features that are strongly correlated to affective state. They reported maximum accuracy of 81.9\% using KNN with 7 features on subject-dependent classification, but they reported the accuracy to go down to 78.9\% for subject independent classification.

Ever since, several novel techniques have been implemented to improve the performance of subject independent classification. Bota et al. \cite{Bota2019} conducted a review of existing machine learning and deep learning techniques for affective state detection from 2001 to 2019. They reported lower performance for subject independent classification as compared to subject dependent classification for most of the
models.

Li et. al \cite{li2020exploring} proposed a technique that assigned variable learnable weights to different physiological signals fed to an attention-based bidirectional LSTM model. They reported an accuracy of 81.1\% for subject-independent classification using the AMIGOS dataset. However, the model's accuracy across other datasets is not known.

Reviews conducted by both Bota et al. \cite{Bota2019} and Xin et al. \cite{Xin} highlighted the lack of generalisation to be a consistent issue with the existing techniques for affective state detection. This has led to several works proposing novel techniques to increase generalisation of affective state detection models. 

Li et al. \cite{li2018exploring} used different automatic feature selection techniques such as Chi-Squared-Based Feature Selection, Mutual Information-Based Feature Selection, ANOVA F-Value-Based Feature Selection, Recursive Feature Elimination (RFE), and L1-Norm Penalty-Based Feature Selection. These techniques were used such that the most important features needed for affective state detection can be extracted and used to train an SVM. They reported an accuracy of 83.3\% for subject independent classification using the SEED dataset and 59.06\% using the DEAP dataset.  

In addition, several domain adaptation techniques are being used in order to select features such that the difference between the distributions of each subject's physiological data is minimized. Chai et al. \cite{chai2016unsupervised} proposed a \textit{Subspace Alignment Autoencoder} to minimize the distribution mismatch of the physiological data of each subject by minimizing the maximum mean discrepancy of reproduced kernel Hilbert space (RKHS). This allowed the alignment of different distributions by mitigating the subject-specific noise from the data. They reported an accuracy 77.8\% using the SEED dataset. 

Based on the literature survey, we identified the potential of decreasing the distance between each subject's distribution of physiological data and use the extracted features to train a more generalised model. However, this entails that a robust method needs to be chosen to calculate the distance between distribution. One such method, Wasserstein Distance, can be retrieved from Optimal Transport Theory (OTT) \cite{monge1781memoir}. Wasserstein Distance allows for the calculation of the distance between distributions while also considering the geometry of the distribution (symmetrical and triangular) \cite{kolouri2018sliced}. Kolouri et al. \cite{kolouri2018sliced} demonstrated the potential of Wasserstein Distance to minimize the distance between an input and target distribution for generative modeling using autoencoders.

Hence, in this paper, we propose a novel loss function using Wasserstein Distance for automatic feature extraction by assigning more weight to features that are closer in distance to each other across all subjects. An autoencoder model is used to generate a latent space that excludes all the features that are influenced by subject-dependent nuances. The goal is to increase the separation between different affective states in the latent space such that a more generalised classification model can be trained.

\section{Methods}
\label{methods}

The datasets are publicly available through their respective organizations, and the code used for the presented work is publicly provided (https://anonymous.4open.science/status/NeurIPS-F252).

\subsection{Optimal Transport Theory}


%
Optimal Transport Theory was first formalized by the French mathematician, Gaspard Monge, in 1781, right before the French revolution \cite{monge1781memoir}. The motivation to study this was to come up with a transport plan to move a mass from one point to another with minimum cost. The outcome of this study formulated a method for measuring the distance between two probability distributions. In the formulation, consider two probability measures $\mu$ and $v$ defined on spaces X and Y respectively. $\mu$ and $v$ have density functions $f$ and $g$ where $d\mu = f(x)dx$ and $dv = g(y)dy$, where $(x, y) \in X$x$Y$. A cost function is defined, $c(x, y)$ which finds the \textit{distance} between a point x and y. The initial choice by Monge was Euclidean distance. A transport plan $T: X \longrightarrow Y$ allows for moving a mass, M, from X to Y such that the measure $\mu$ of the mass in space X is the same as the measure $v$ of the mass in space Y after the mass is transported from X to Y:

\begin{equation}
    \mu[M] = v[T(M)]
\end{equation}

Once a plan T has been found, the cost associated with it is: 

\begin{equation}
    I(T,f,g,c) = \int_X c(x, T(x))f(x)dx \footnote{The equation is adapted from \cite{yang2018application}}
\end{equation}

While there can be many solution to T, the aim is to minimize the overall cost

\begin{equation}
    I(f,g,c) = \inf_{T \in M} \int_X c(x, T(x))f(x)dx, \footnote{ibid.}
\end{equation}

where $M$ is the set of all transport plans T that transfers $f$ to $g$. In the formulation, the optimal cost of transporting $x$ to $y$ is also known as the Wasserstein Distance.

\begin{equation}
    W_p(\mu, v) = ( \inf_{T \in M} \int_{\mathbb{R}^n} |x - T(x)|^p d\mu(x) )^{\frac{1}{p}}, \mu, v \in \mathcal{P}_p(X),\footnote{ibid} 
\end{equation}

where $\mathcal{P}_p(X)$ is the set of probability measures with finite moments of order $p$ \cite{yang2018application}. In this paper, we will focus on using the Wasserstein Distance to measure the distance between a subjects' individual distribution and the group's distribution. 



\begin{figure}
\centering
  \includegraphics[scale=0.02]{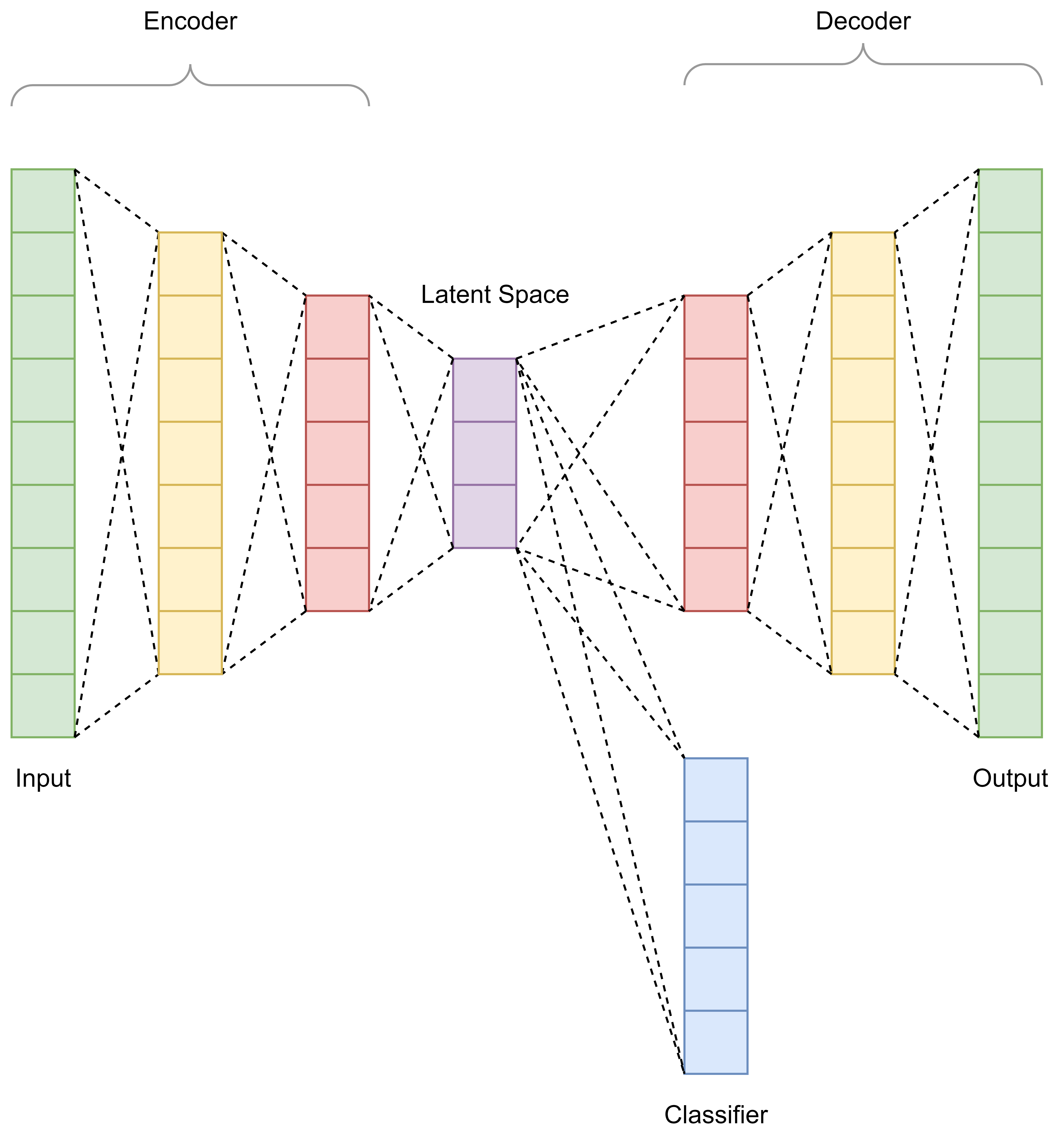}
  \caption{Architecture of the Autoencoder model with a classifier attached to the latent space.}
  \label{fig:autoencoder}
\end{figure}

\subsection{Models}

In this work, we focus on using an autoencoder architecture to reduce noise along with our novel cost function through dimensionality reduction. In addition to a reconstruction loss, a classifier is attached to the latent space to ensure it is discriminant. Figure \ref{fig:autoencoder} shows the architecture of our model, and both components (encoder and classifier) are trained simultaneously. The proposed cost function is not restricted to the autoencoder model and can be used with any suitable machine learning algorithm. The following section will address how the loss function can be integrated in the autoencoder architecture. We use a standard autoencoder model with Mean Squared Error (MSE) for a baseline to compare it with our novel cost function. 

\subsection{Autoencoder Error Decomposition}

The group reconstruction error, $r_{g}$, of the autoencoder for the subject-independent model is given as

\begin{equation}
    r_{g} = \frac{1}{N} \sum\limits_{n=1}^N \mathcal{L}_a(a; g_{n,x}, g^{\prime}_{n,x})
\label{recon_group}
\end{equation}

where $N$ is the total number of samples across all subjects, $\mathcal{L}_a$ is any loss function for reconstruction (Mean-Squared Error), and $a$ is the encoder and decoder architecture of the network. The reconstruction error will be used in conjunction with the classifier loss. The classifier head, attached to the latent space, is used to calculate the group classifier error and subject specific classifier error. The group classifier uses all labeled samples available across all subjects.

\begin{equation}
    c_{g} = \frac{1}{N} \sum\limits_{n=1}^N \mathcal{L}_c(c; g_{n,x}, g_{n,y}) * \lambda_g
\end{equation}

where $\mathcal{L}_c$ is any classification loss function (Categorical Cross-Entropy) and $\lambda_g$ is a regularizer term used to scale the importance of the group classifier error during backpropagation. The definition of $\lambda_g$ is given in equation \ref{eq:lamdag}. Subject $i$'s specific loss is computed using all labeled samples available $S_{i}$ and any loss function $\mathcal{L}_c$. 

\begin{equation}
    c_{s,i} = \frac{1}{N_{i}} \sum\limits_{n=1}^{N_{i}} \mathcal{L}_c(f; s_{n,x}, s_{n,y}) * \lambda_{s,i}
\end{equation}

$\lambda_{s,i}$ is used to scale the importance of a specific subject in the calculation of the overall loss for backpropogation. The overall classification loss is given as: 

\begin{equation}
    c_s = \sum \limits_{i=i}^{S} c_{s,i}
\end{equation}

where $S$ is the number of subjects. Optimal Transport Theory, specifically, Wasserstein Distance, is used to calculate $\lambda_{s,i}$ by measuring the distance between the group distribution and subject distribution. Equations \ref{eq:alphas} and \ref{eq:labdas} denote the formulation for $\lambda_{s,i}$.

\begin{equation}
    \alpha_{s,i} = EMD(G, S_{i}) = \inf_{\gamma \in \phi (G, S_{i})} \mathbb{E}_{(G, S_{i}) \sim \gamma} [|| x - y ||]
    \label{eq:alphas}
\end{equation}

$\alpha_{s,i}$ is normalized and subtracted from 1 to ensure that the further away a subject $i$ is from the group, the smaller the scaling factor, $\lambda_{s,i}$, is.

\begin{equation}
    \lambda_{s,i} = 1 - \frac{\alpha_{s,i}}{|    \sum\limits_{i = 1}^{S} \alpha_{s,i} |}
    \label{eq:labdas}
\end{equation}

Finally, $\lambda_g$ is determined by ensuring that the regularizer terms sum to 1. 

\begin{equation}
    \lambda_g + \sum\limits_{i=1}^S \lambda_{s,i} = 1
    \label{eq:lamdag}
\end{equation}

\subsection{Datasets}
Experiments in this work were conducted on popular, public physiological datasets: WESAD, AMIGOS, CLAS, and DEAP.

WESAD is a multimodal dataset for wearable stress and affect detection with data recorded using a wrist-worn device (sensors: PPG, accelerometer, electrodermal activity, and body temperature) and a chest-worn device (sensors: ECG, accelerometer, EMG, respiration, and body temperature). The dataset consists of 15 subjects (aged 24–35 years) with 100 min of data each. The dataset was recorded with the goal of detecting and distinguishing between three affective states: neutral, stress, amusement which were denoted using self-reports for subjective experience during the emotional stimulus along with the study protocol of inducing emotion \cite{schmidt2018introducing}. 

The AMIGOS dataset consists of data from sensors (EEG, ECG and GSR), full-body videos, and depth videos from 40 volunteers. Data was collected while watching 16 short videos and 37 of the 40 volunteers watching 4 long-videos. Both self-assessment (valence-arousal, control, familiarity, like-dislike, and selection of basic emotions) and external assessment of participants’ levels of valence were used to annotate the dataset. The participants were also asked to fill forms with Personality Traits and Positive and Negative Affect Schedule (PANAS) questionnaires \cite{miranda2018amigos}. 

DEAP contains EEG, GSR, RESP, SKT, EMG, EOG, BVP data from 32 volunteers while they were watching 40 one minute-long music videos. Additionally, frontal face video was recorded for 22 of the participants. The dataset was self-annotated after each video with the following labels: arousal, valence, like-dislike, familiarity, and dominance by the volunteers \cite{koelstra2011deap}.

PPG, EEG and GSR data from 60 patients engaged in cognitively difficult tasks from was collected in the CLAS dataset. The cognitively difficult tasks used in this dataset is: Stroop test, math test, logic problem test, and emotionally evoking stimuli \cite{markova2019clas}. The dataset was labeled based on the study protocol (i.e., high cognitive load during logic and math problems and low cognitive load during neutral stimuli). 

\section{Results}
\label{results}

The results presented in this work use a standard 10-fold Leave-One-Subject-Out (LOSO) where the models were trained on all subjects excluding one and testing on the excluded subject.

\begin{figure*}
    \centering
    \subfloat[\centering WESAD MSE]{{\includegraphics[trim= 0 0 0 20,clip,width=0.20\textwidth]{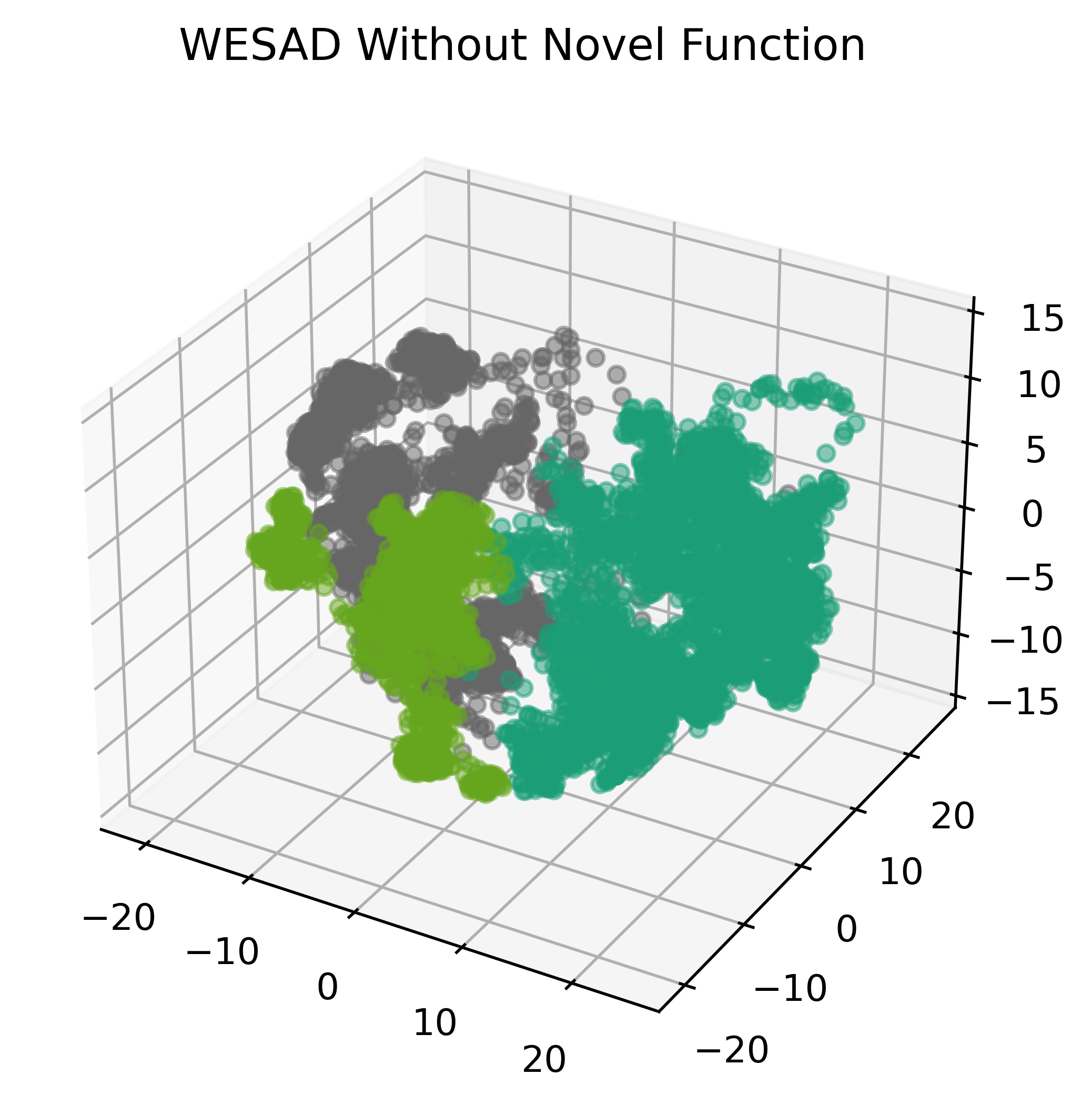} }}%
    \qquad
    \subfloat[\centering DEAP MSE]{{\includegraphics[trim= 0 0 0 20,clip,width=0.20\textwidth]{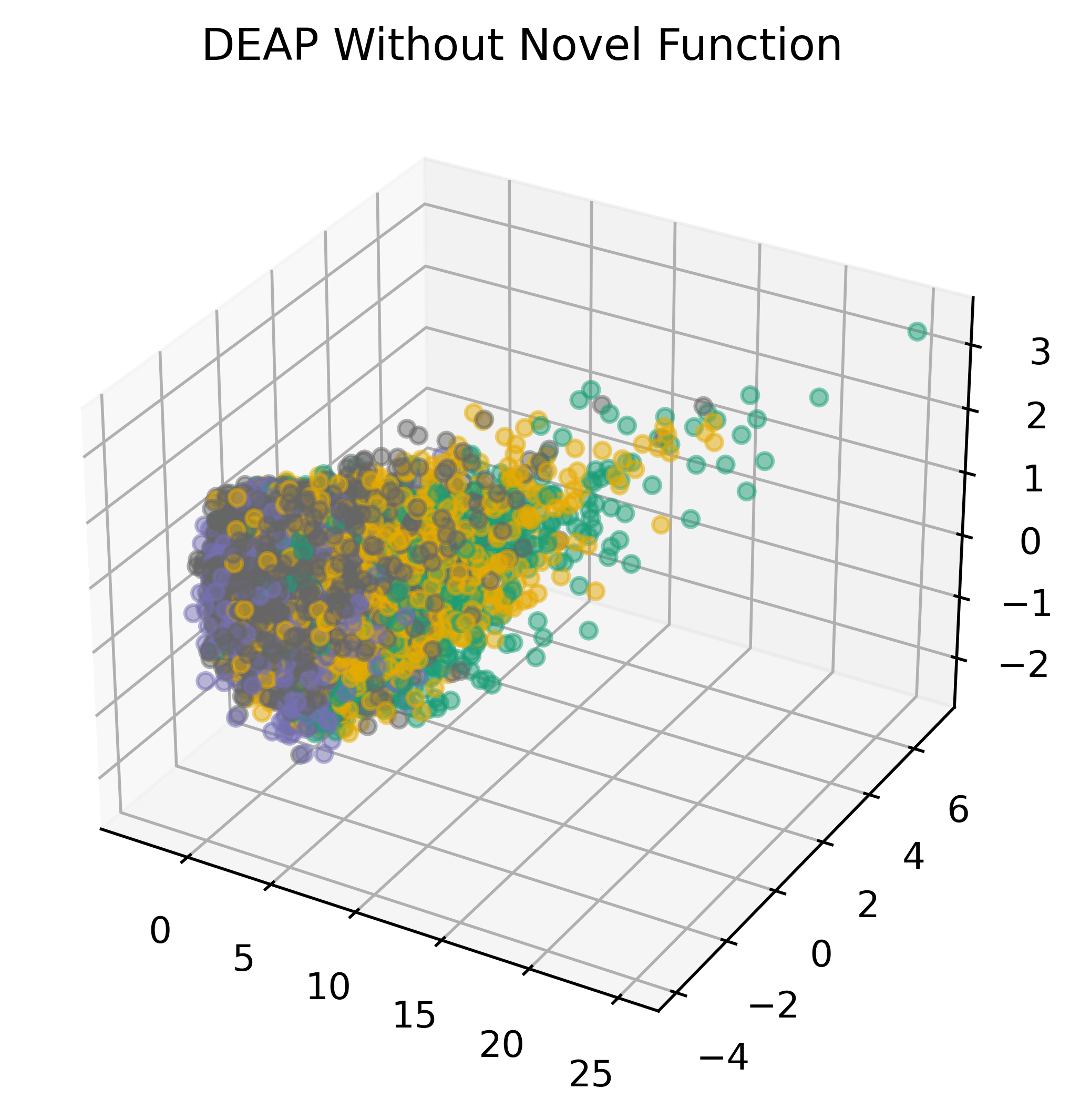} }}%
    \qquad
    \subfloat[\centering CLAS MSE]{{\includegraphics[trim= 0 0 0 20,clip,width=0.20\textwidth]{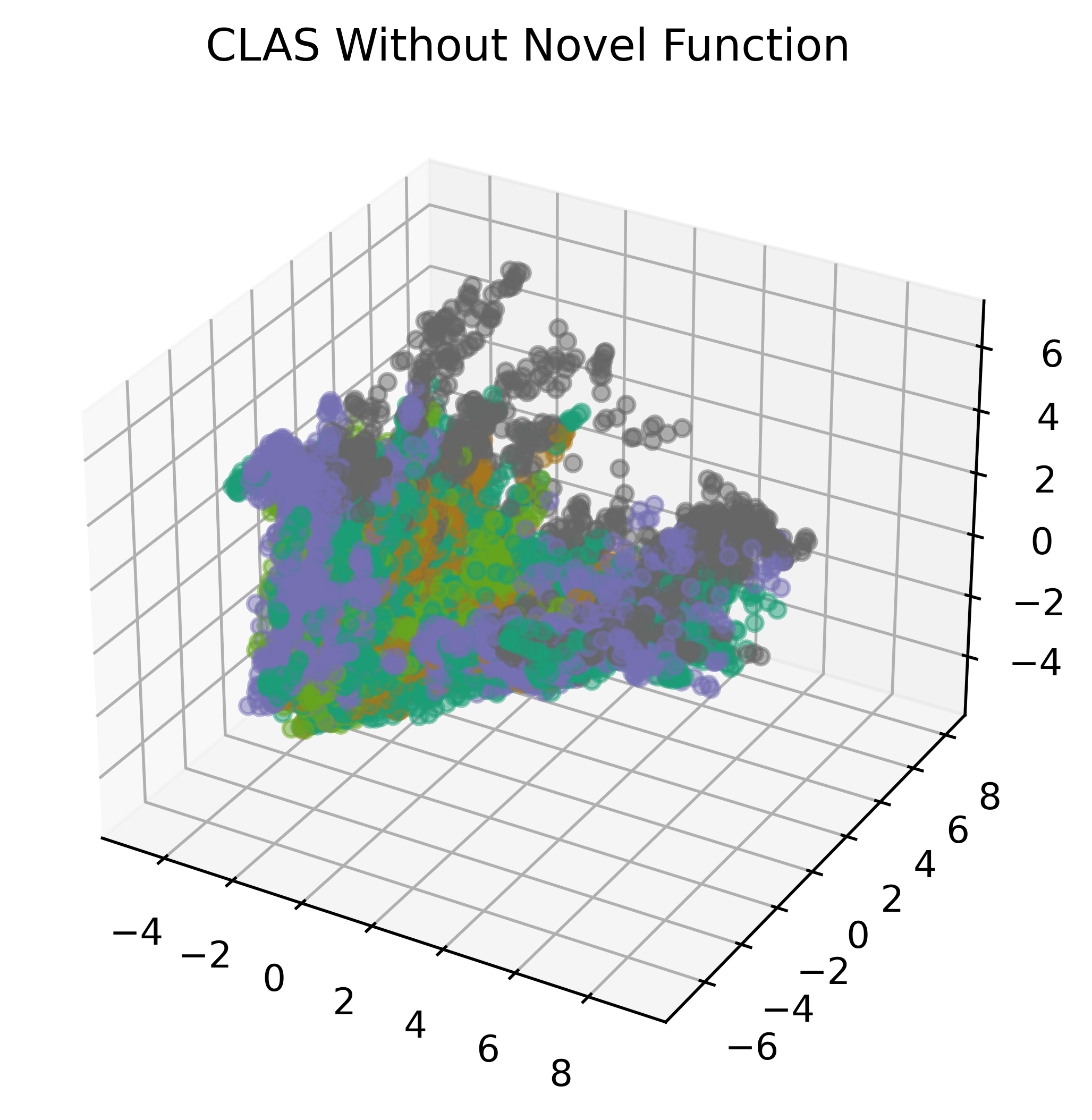} }}%
    \qquad
    \subfloat[\centering AMIGOS MSE]{{\includegraphics[trim= 0 0 0 20,clip,width=0.20\textwidth]{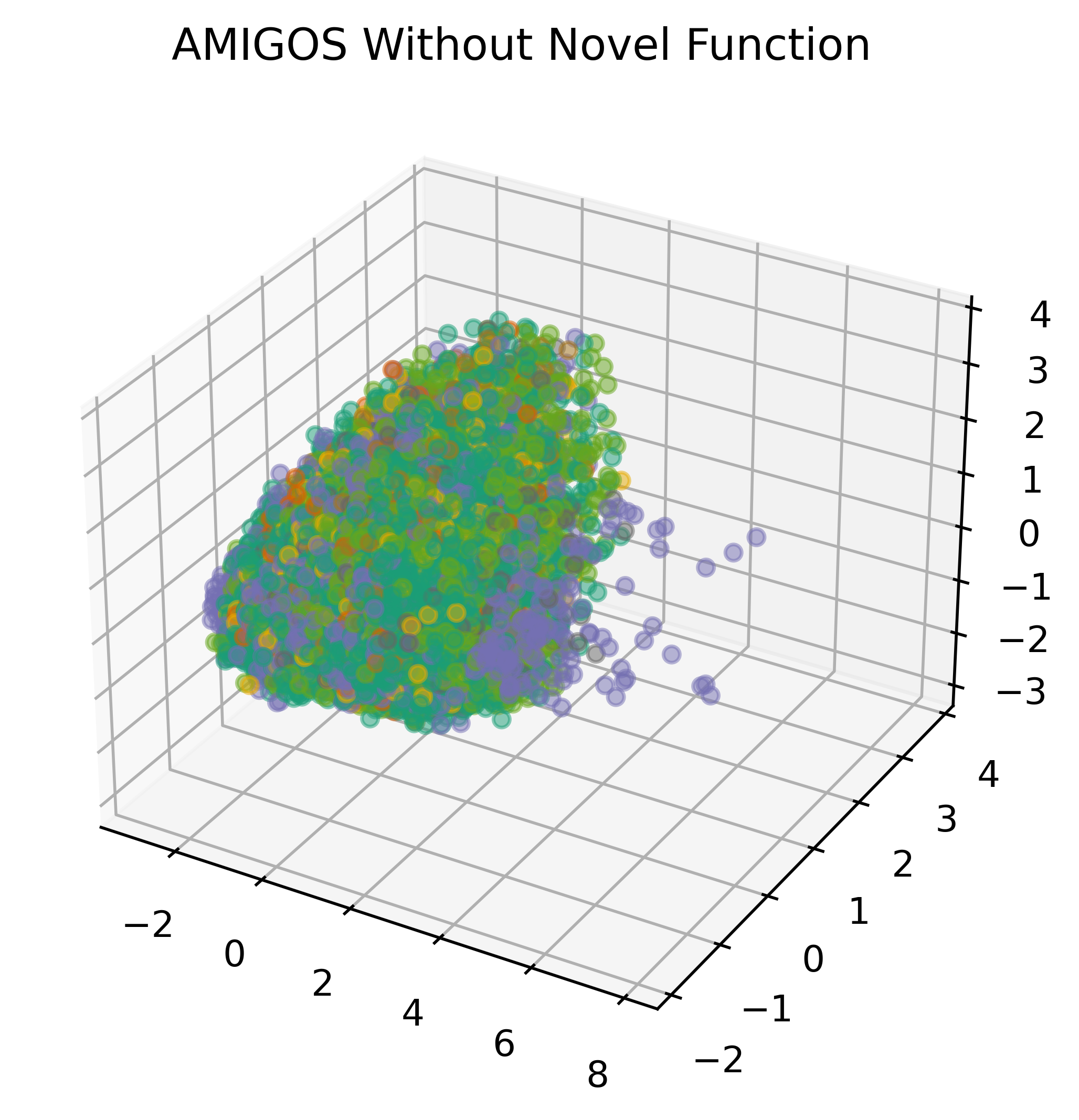} }}%
    
    \vskip\baselineskip
    
    \subfloat[\centering WESAD Custom]{{\includegraphics[trim= 0 0 0 20,clip,width=0.20\textwidth]{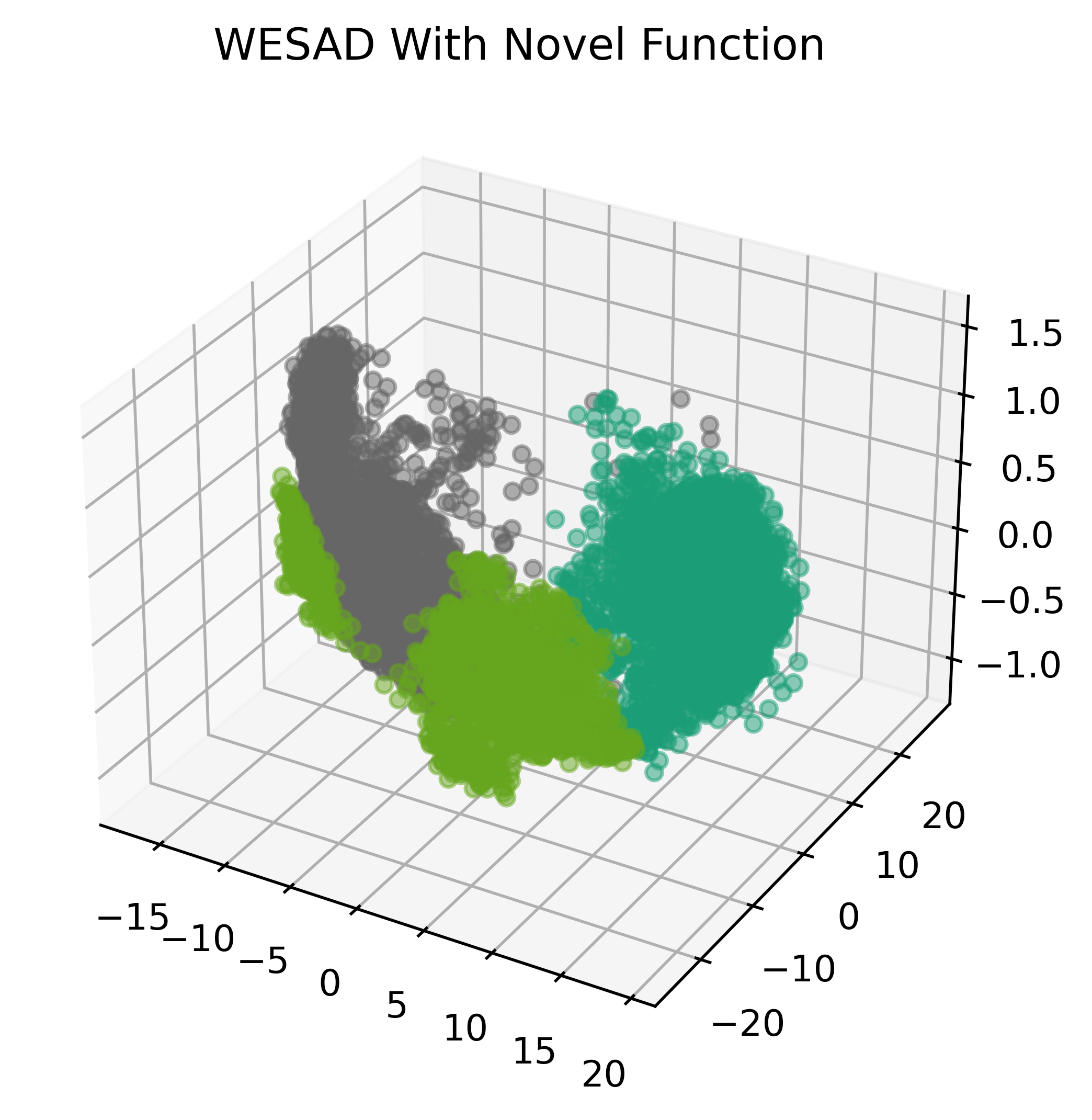} }}%
    \qquad
    \subfloat[\centering DEAP Custom]{{\includegraphics[trim= 0 0 0 20,clip,width=0.20\textwidth]{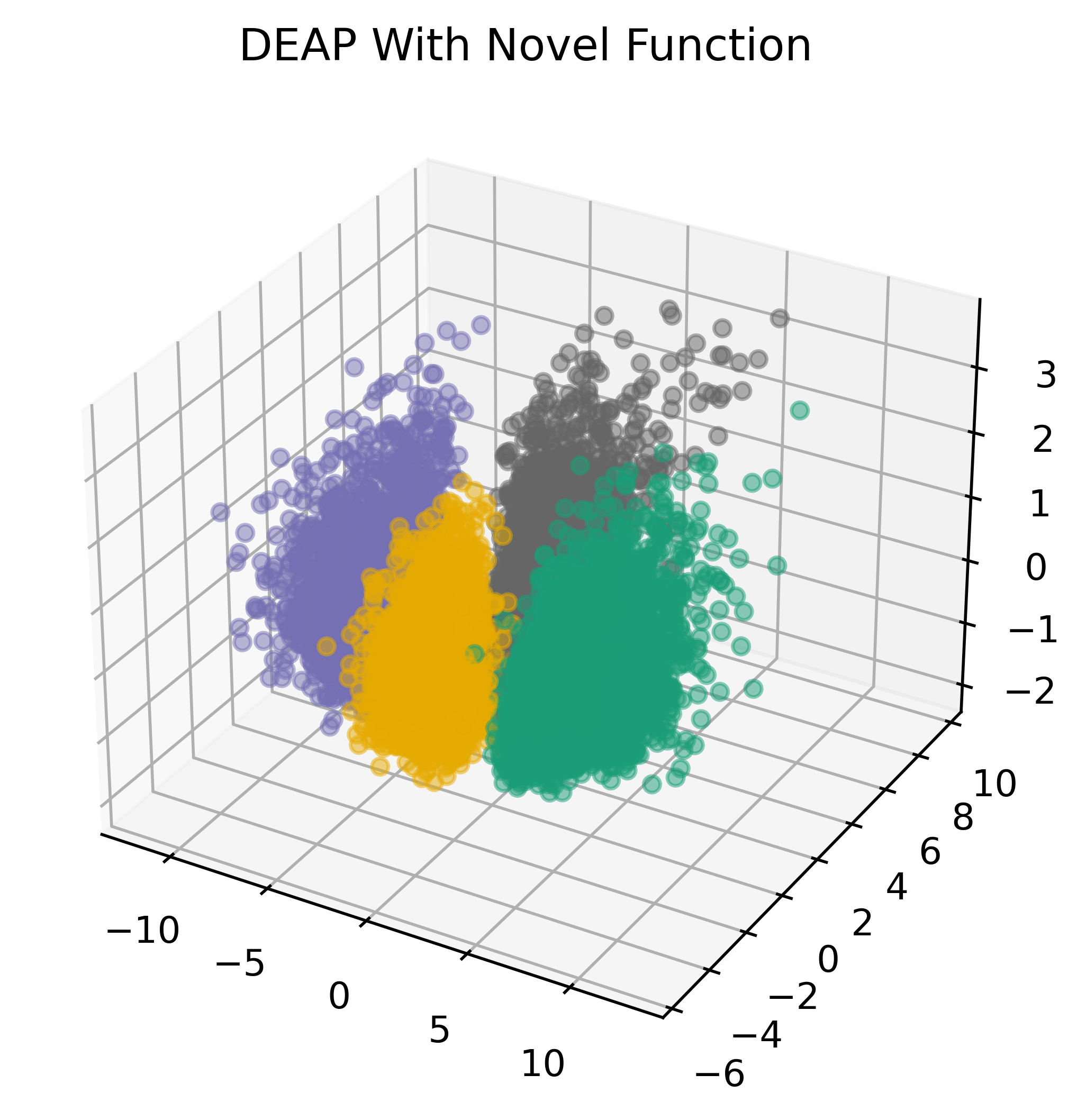} }}%
    \qquad
    \subfloat[\centering CLAS Custom]{{\includegraphics[trim= 0 0 0 20,clip,width=0.20\textwidth]{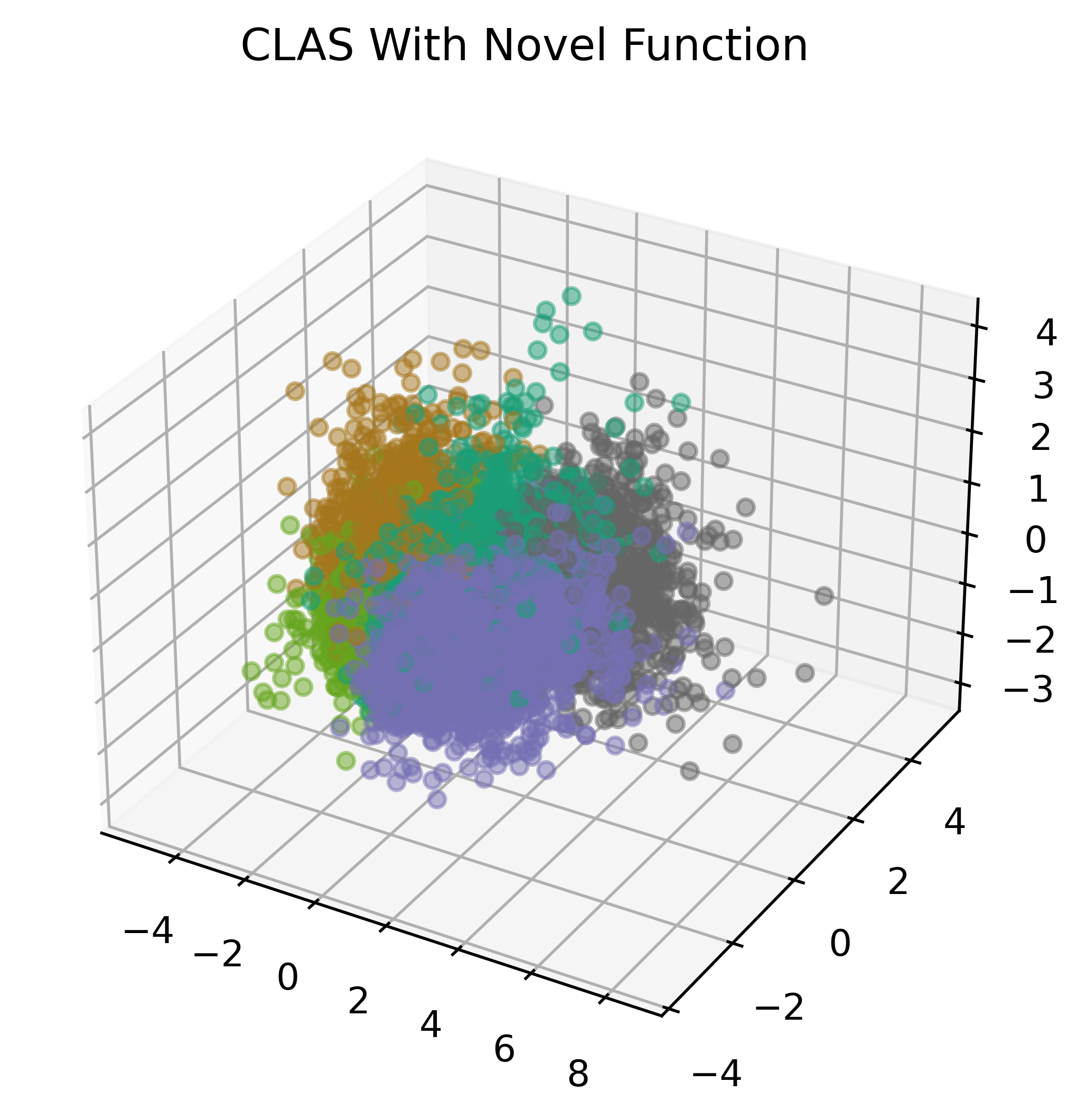} }}%
    \qquad
    \subfloat[\centering AMIGOS Custom]{{\includegraphics[trim= 0 0 0 20,clip,width=0.20\textwidth]{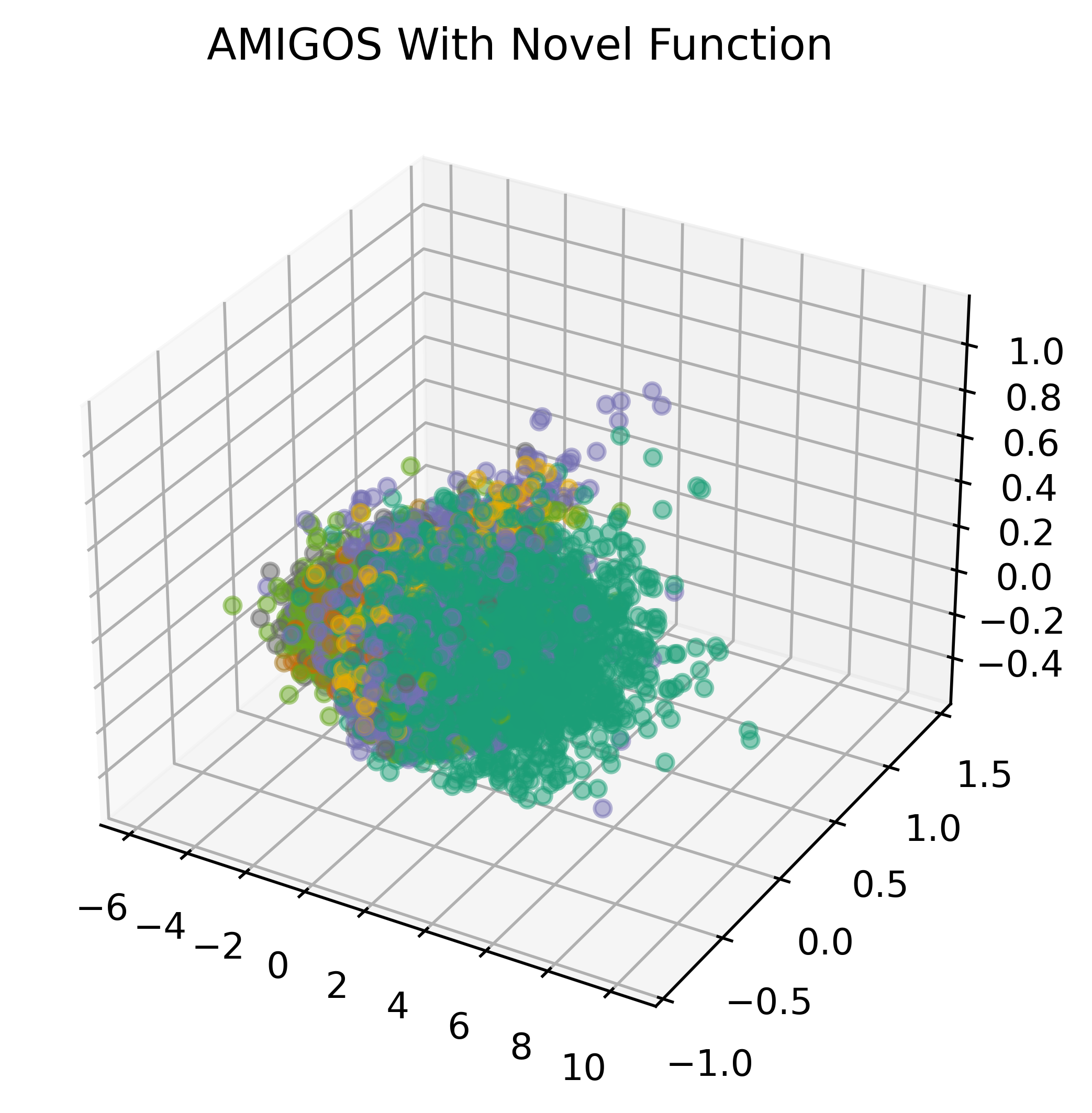} }}%

    \caption{Autoencoder latent space visualization using PCA} 
    \label{fig:pca}
\end{figure*}

The dimensionality of the latent space is too high to be visualized, so Principal Component Analysis (PCA) is used to condense the space into three components. Figure \ref{fig:pca} compares the latent space using MSE with our cost function. Figure \ref{fig:pca}(a) - \ref{fig:pca}(d) shows a large overlap of classes which would lead to imprecise decision boundaries. Specifically consider Figure \ref{fig:pca}(b) where all classes are highly interrelated. In contrast, the latent space in \ref{fig:pca}(f) shows a clear separation of classes by utilizing Wasserstein Distance. While the advantages of the novel cost function is distinct in three dimensions, the latent space has higher dimensionality, and the improvements are expressed using the euclidean distance between centroids of classes and the euclidean distance between closest points between different classes.

\begin{figure*}
    \centering
    \subfloat[\centering WESAD]{{\includegraphics[width=0.49\textwidth]{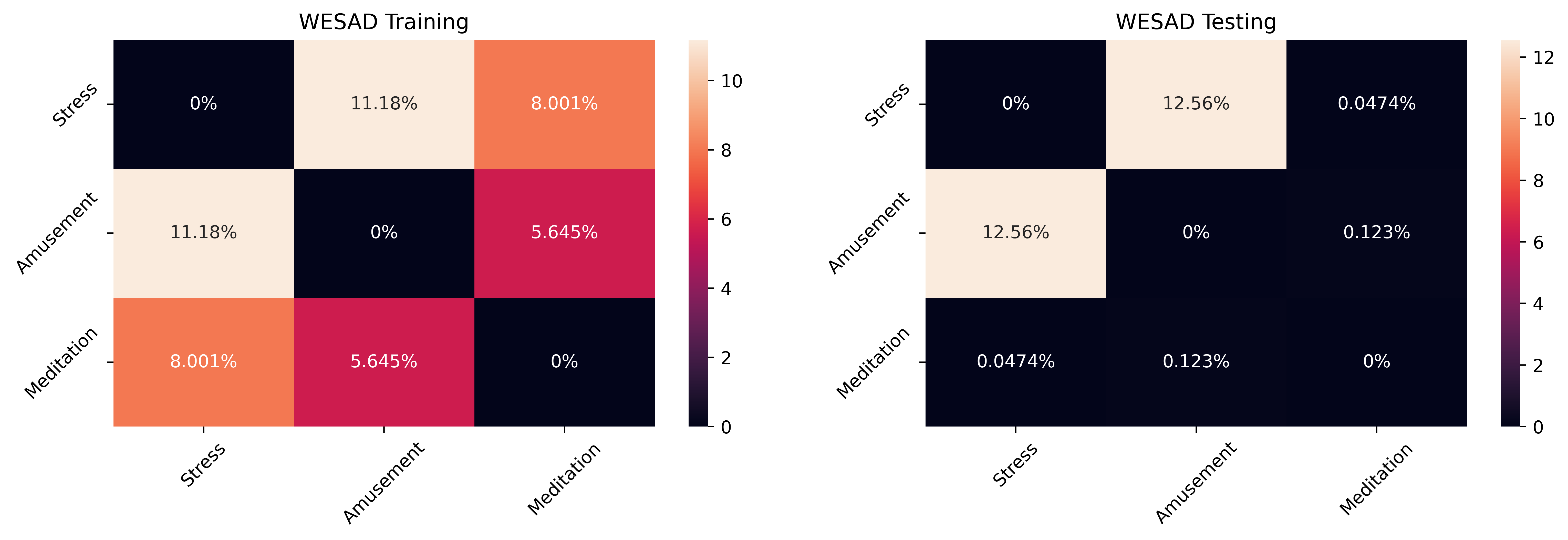} }}%
    \subfloat[\centering DEAP]{{\includegraphics[width=0.49\textwidth]{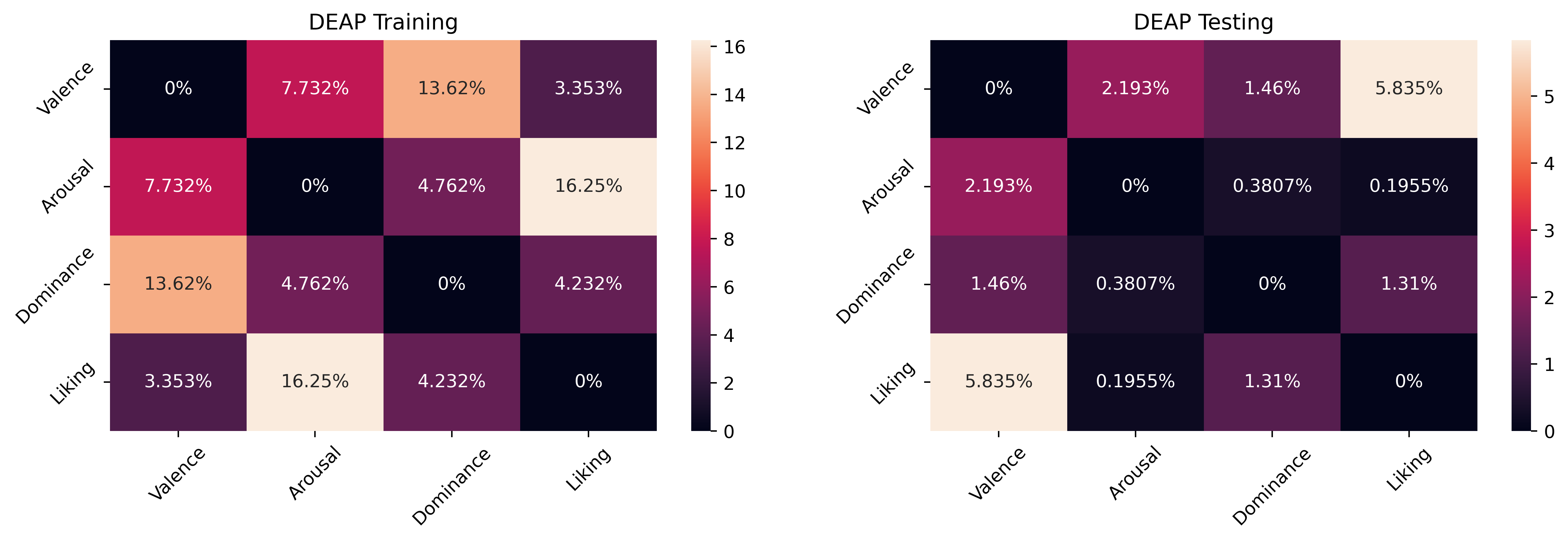} }}%
    \vskip\baselineskip
    \subfloat[\centering CLAS]{{\includegraphics[width=0.49\textwidth]{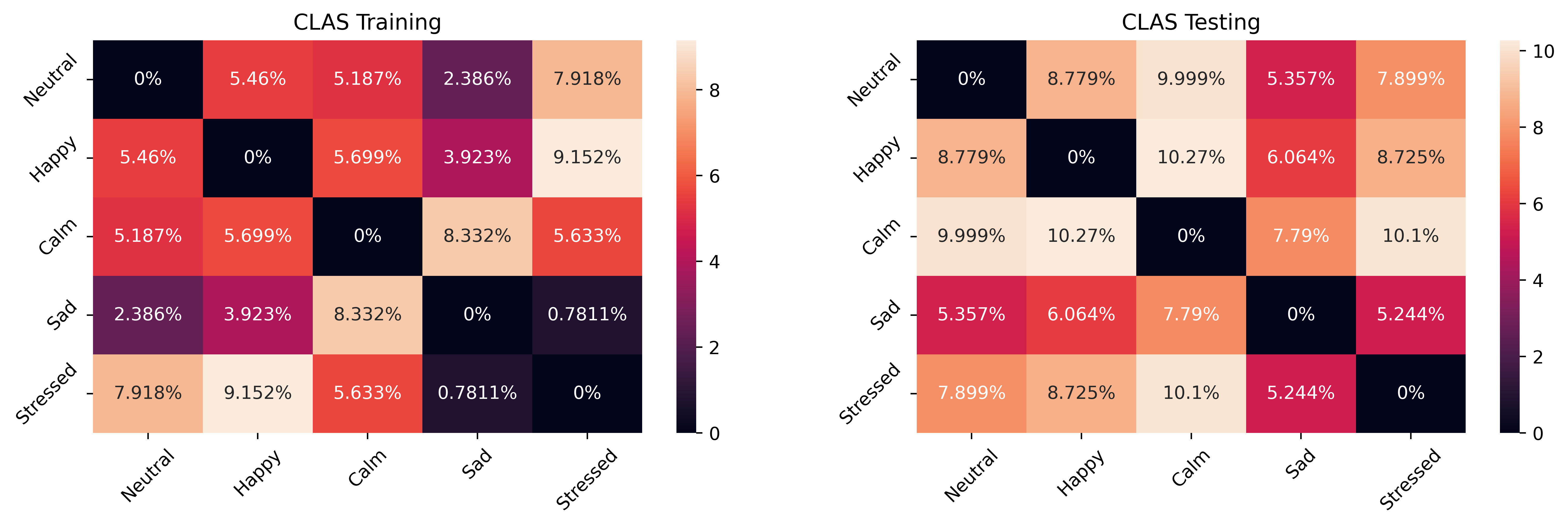} }}%
    \subfloat[\centering AMIGOS]{{\includegraphics[width=0.49\textwidth]{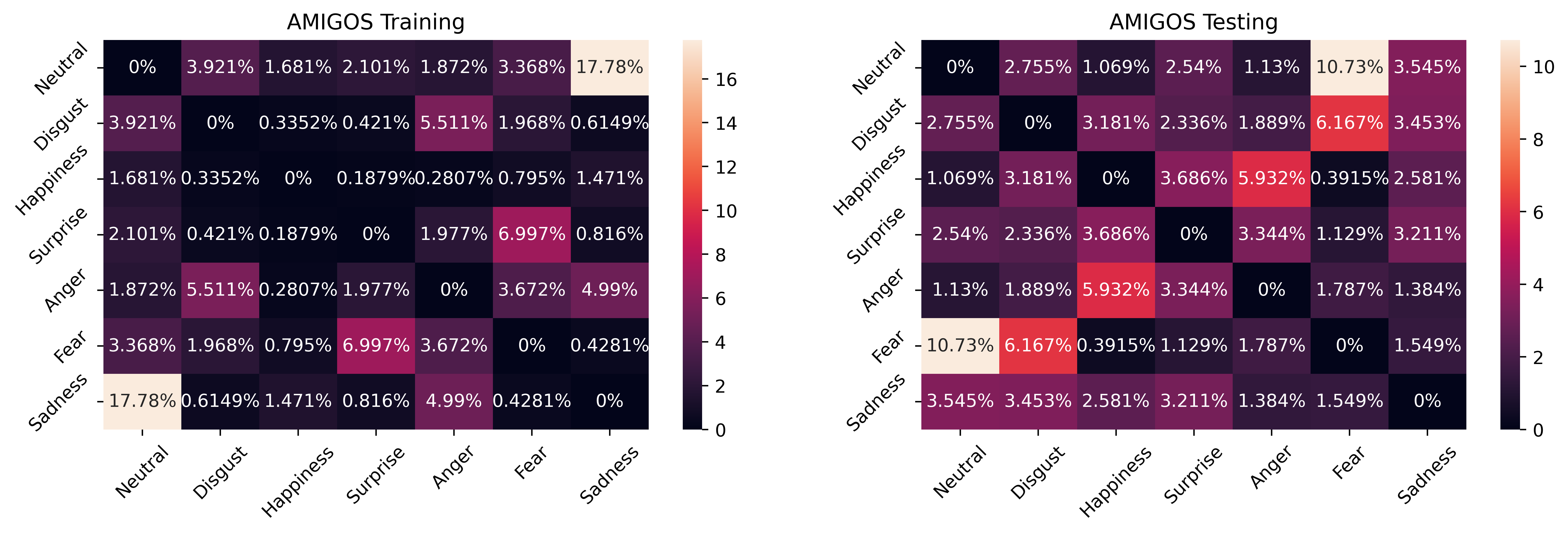} }}%
    
    \caption{Distance increase when using the novel cost function over MSE expressed as a percentage for training and testing data (LOSO) for class centroids.} 
    \label{fig:percent_change_centroids}
\end{figure*}

\begin{figure*}
    \centering
    \subfloat[\centering WESAD]{{\includegraphics[width=0.49\textwidth]{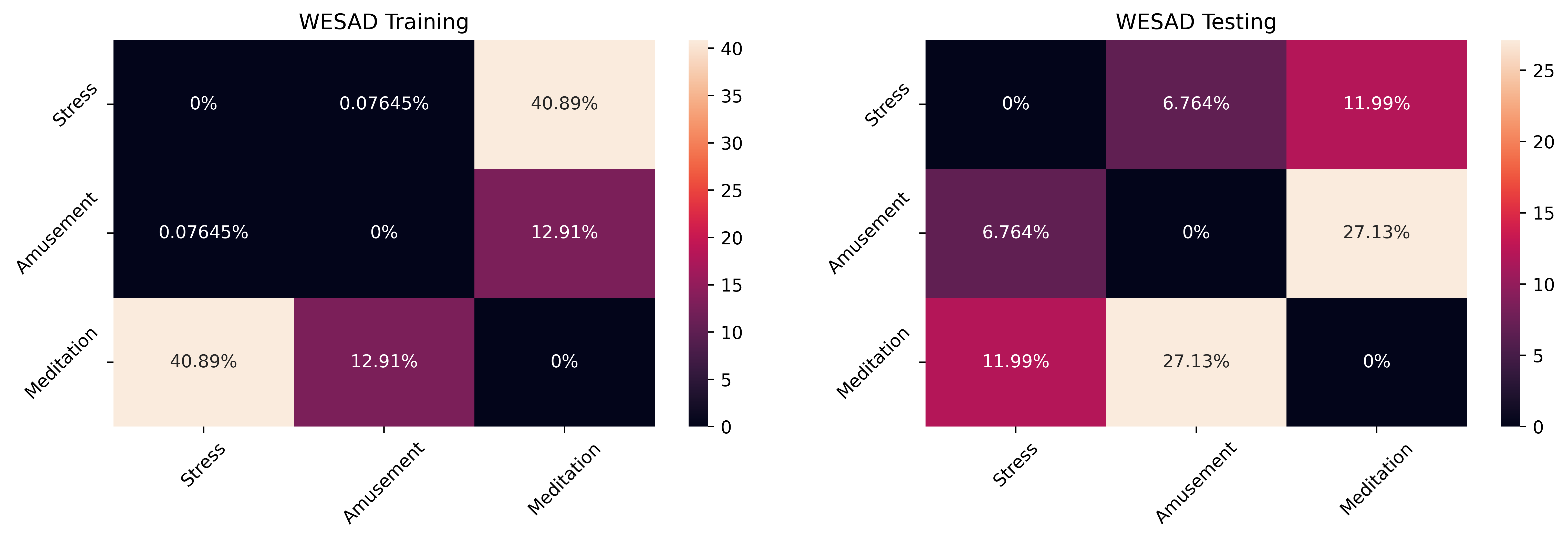} }}%
    \subfloat[\centering DEAP]{{\includegraphics[width=0.49\textwidth]{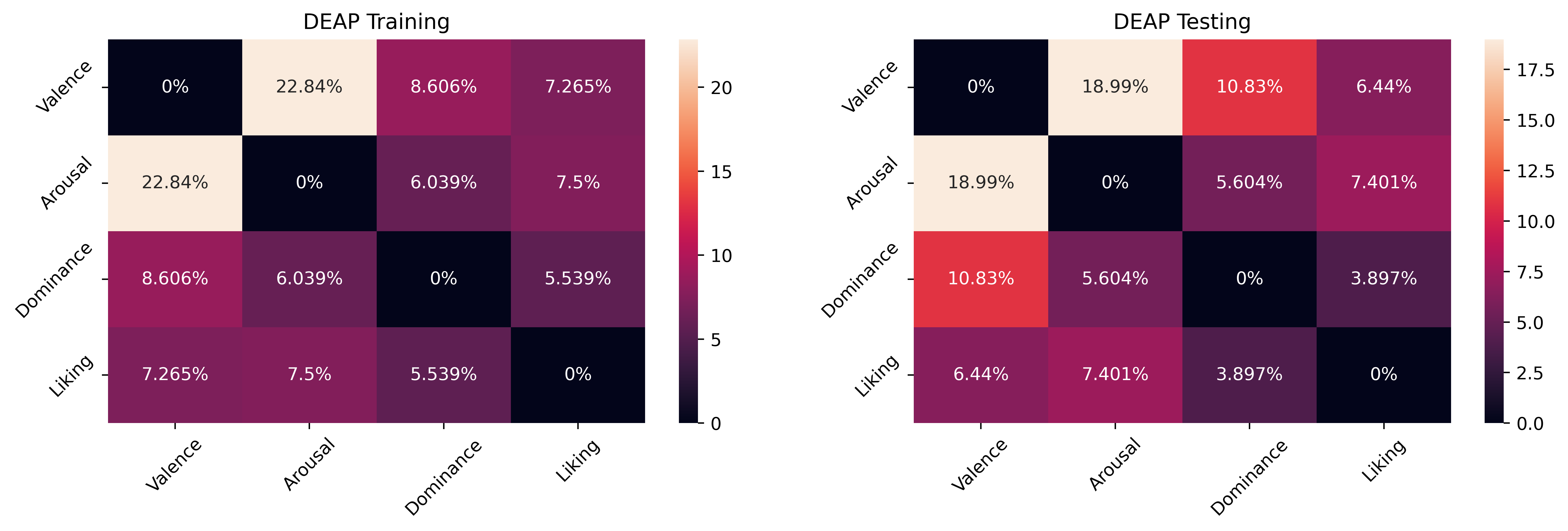} }}%
    \vskip\baselineskip
    
    \subfloat[\centering CLAS]{{\includegraphics[width=0.49\textwidth]{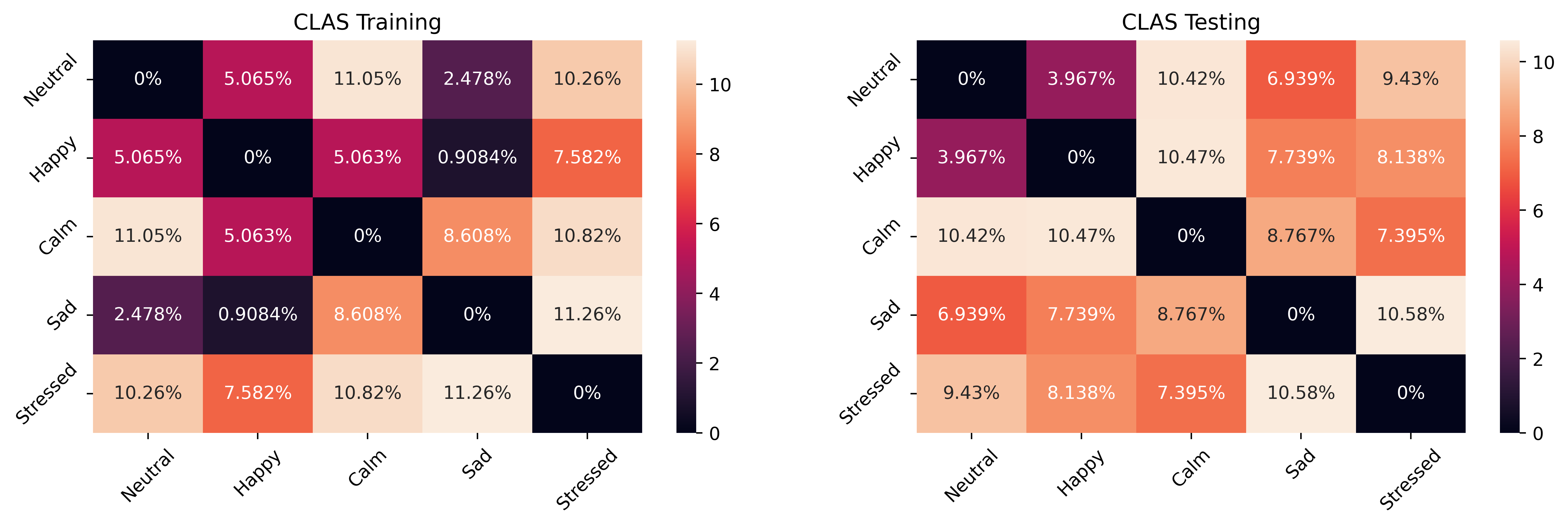} }}%
    \subfloat[\centering AMIGOS]{{\includegraphics[width=0.49\textwidth]{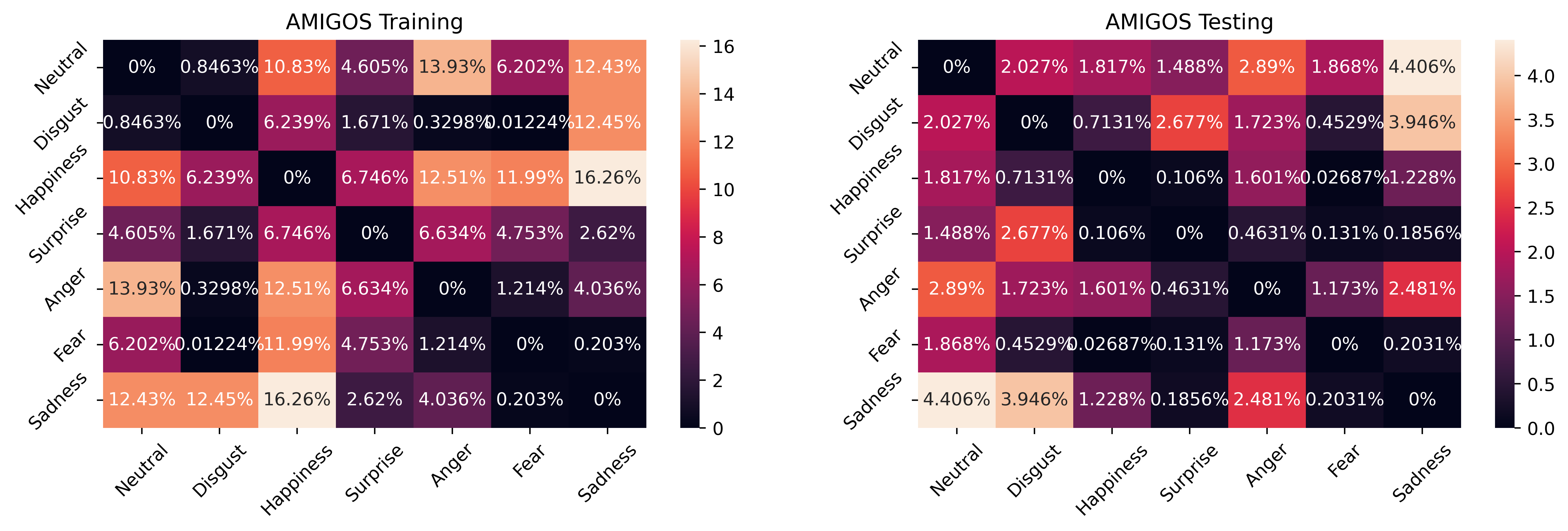} }}%

    \caption{Distance increase when using the novel cost function over MSE expressed as a percentage for training and testing data (LOSO) for the minimum distance between classes.} 
    \label{fig:percent_change_minmax}
\end{figure*}

Figure \ref{fig:percent_change_centroids} highlights the percent increase between the centroids of classes in all datasets with the novel cost function for training and testing data. For all datasets, there is a significant increase in the distance over MSE. Additionally, Figure \ref{fig:percent_change_minmax} shows the increase in the minimum distance between samples of classes.

The accuracy of our models is presented in Table \ref{tab:acc} through a comparison of MSE with our novel cost function using LOSO. The table shows that the proposed cost function is able to achieve better performance on most datasets and performed the same as the MSE in the worst case scenario. The experiments presented in this work were conducted on a computer with a AMD Ryzen 9 5900HX, NVIDIA GeForce RTX 3070, and 32 GB of RAM with an approximate run time of 45 minutes per dataset.

\begin{table}[]
\centering
\caption{A comparison of the accuracy between using the standard MSE for reconstruction loss and the novel cost function.}
\begin{tabular}{lll}
\hline
       & MSE Accuracy & Novel Cost Accuracy \\ \hline
WESAD  & 80.31\% $\pm$ 1.13\%     & 84.11\% $\pm$ 1.05\%             \\
CLAS   & 77.23\% $\pm$ 0.76\%     & 79.63\% $\pm$ 0.86\%            \\
AMIGOS & 79.56\% $\pm$ 1.44\%     & 79.97\% $\pm$ 1.75\%            \\
DEAP   & 66.22\% $\pm$ 1.87\%     & 71.64\% $\pm$ 1.81\%            \\ \hline
\end{tabular}
\label{tab:acc}
\end{table}

\section{Conclusion and Future Works}
\label{conclusion}
The lack of generalizability of affective state detection models has been a consistent issue. It is often due to a large variation in the distribution of physiological data for a set of individuals in reaction to the same stimuli. This leads to subject dependent noise that affects the performance of a model. Our approach employs autoencoders that can inherently reduce noise and extract relevant features through dimensionality reduction. In order to further reduce the subject-dependent noise, we introduce a novel cost function that uses Optimal Transport Theory to lower the importance of uncommon patterns across individuals. This results in a latent space of affective state classes with increased separability.

The model was trained and tested on four different datasets and the performance of the proposed cost function was compared to the state-of-the-art MSE. From our study, we show the proposed cost function significantly increases the distance between classes in the latent space across all datasets. An average increase of $14.75\%$ and $17.75\%$ (from benchmark to proposed loss function) was found for minimum distance between classes and centroid euclidean distance respectively.

Additionally, our proposed cost function increases the robustness of affective computing models allowing for generalizability as the population increases. The larger the population, the more diverse the physiological responses to stimuli. This results in overlap between class clusters leaving no space to draw accurate decision boundaries. Since state-of-the-art models do not have a mechanism to accommodate this increased noise, their performance might decreases as the population size increase. Alternatively, our proposed loss function mitigates the affect of increased subject-dependent noise while increasing the performance by allowing the model to train on more data.

However, commonly used public physiological datasets do not contain samples from a large population as the compilation of a large dataset is expensive. In our experiments, we used datasets with a smaller population size and were still able to show an increase in the distance between classes. Even though there was a significant increase in the separation between the classes, the accuracy of a multilayer perceptron trained on the latent space was nearly equivalent to that of the state-of-the-art models (see Table \ref{tab:acc}). We were not able to highlight the full extent of the cost function due to the lack of a large scale physiological dataset. Future works include implementation of more complex encoder and decoder models (for e.g., LSTMs and CNNs), exploration of different classifiers on the latent space in order to further increase the accuracy, and investigation of the cost function on a large scale dataset.

\bibliographystyle{unsrt}  
\bibliography{references}

\end{document}